\documentclass[conference]{IEEEtran}
\IEEEoverridecommandlockouts
\usepackage{cite}
\usepackage{amsmath,amssymb,amsfonts}
\usepackage{algorithmic}
\usepackage{graphicx}
\usepackage{textcomp}
\usepackage{xcolor}
\usepackage{siunitx} 
\usepackage[labelformat=simple, subrefformat=simple]{subcaption}
\usepackage{flushend} 

\makeatletter 
\newcommand{\linebreakand}{%
  \end{@IEEEauthorhalign}
  \hfill\mbox{}\par
  \mbox{}\hfill\begin{@IEEEauthorhalign}
}
\makeatother 

\def\BibTeX{{\rm B\kern-.05em{\sc i\kern-.025em b}\kern-.08em
    T\kern-.1667em\lower.7ex\hbox{E}\kern-.125emX}}
\begin{document}

\title{
A Sequential Hermaphrodite Coupling Mechanism for Lattice-Based Modular Robots
}

\author{
    \IEEEauthorblockN{1\textsuperscript{st} Keigo Torii\textsuperscript{*}}
    \IEEEauthorblockA{
        \textit{Space Robotics Lab. (SRL),}\\
        \textit{Department of Aerospace Engineering,}\\
        \textit{Tohoku University}\\
        Sendai, Japan \\
        \small{\tt torii.keigo.p1@dc.tohoku.ac.jp}\\
        *Corresponding author\\
    }
    \and
    \IEEEauthorblockN{2\textsuperscript{nd} Kentaro Uno}
    \IEEEauthorblockA{
        \textit{Space Robotics Lab. (SRL),}\\
        \textit{Department of Aerospace Engineering,}\\
        \textit{Tohoku University}\\
        Sendai, Japan \\
        \small{\tt unoken@tohoku.ac.jp}
    }
    \and
    \IEEEauthorblockN{3\textsuperscript{rd} Shreya Santra}
    \IEEEauthorblockA{
        \textit{Space Robotics Lab. (SRL),}\\
        \textit{Department of Aerospace Engineering,}\\
        \textit{Tohoku University}\\
        Sendai, Japan \\
        \small{\tt shreya.santra@tohoku.ac.jp}
    }
    \and
    \linebreakand
    \IEEEauthorblockN{4\textsuperscript{th} Kazuya Yoshida}
    \centering
    \IEEEauthorblockA{
        \textit{Space Robotics Lab. (SRL),}\\
        \textit{Department of Aerospace Engineering,}\\
        \textit{Tohoku University}\\
        Sendai, Japan \\
        \small{\tt yoshida.astro@tohoku.ac.jp}
    }
}

\maketitle
\begin{abstract}
Lattice-based modular robot systems are envisioned for large-scale construction in extreme environments, such as space.
Coupling mechanisms for heterogeneous structural modules should meet all of the following requirements: single-sided coupling and decoupling, flat surfaces when uncoupled, and coupling to passive coupling interfaces as well as coupling behavior between coupling mechanisms.
The design requirements for such a coupling mechanism are complex.
We propose a novel shape-matching mechanical coupling mechanism that satisfies these design requirements.
This mechanism enables controlled, sequential transitions between male and female states.
When uncoupled, all mechanisms are in the female state.
To enable single-sided coupling, one side of the mechanisms switches to the male state during the coupling process.
Single-sided decoupling is possible not only from the male side but also from the female side by forcibly switching the opposite mechanism's male state to the female state.
This coupling mechanism can be applied to various modular robot systems and robot arm tool changers.
\end{abstract}


\begin{figure}[t]
    \centering
    \includegraphics[scale=0.089]{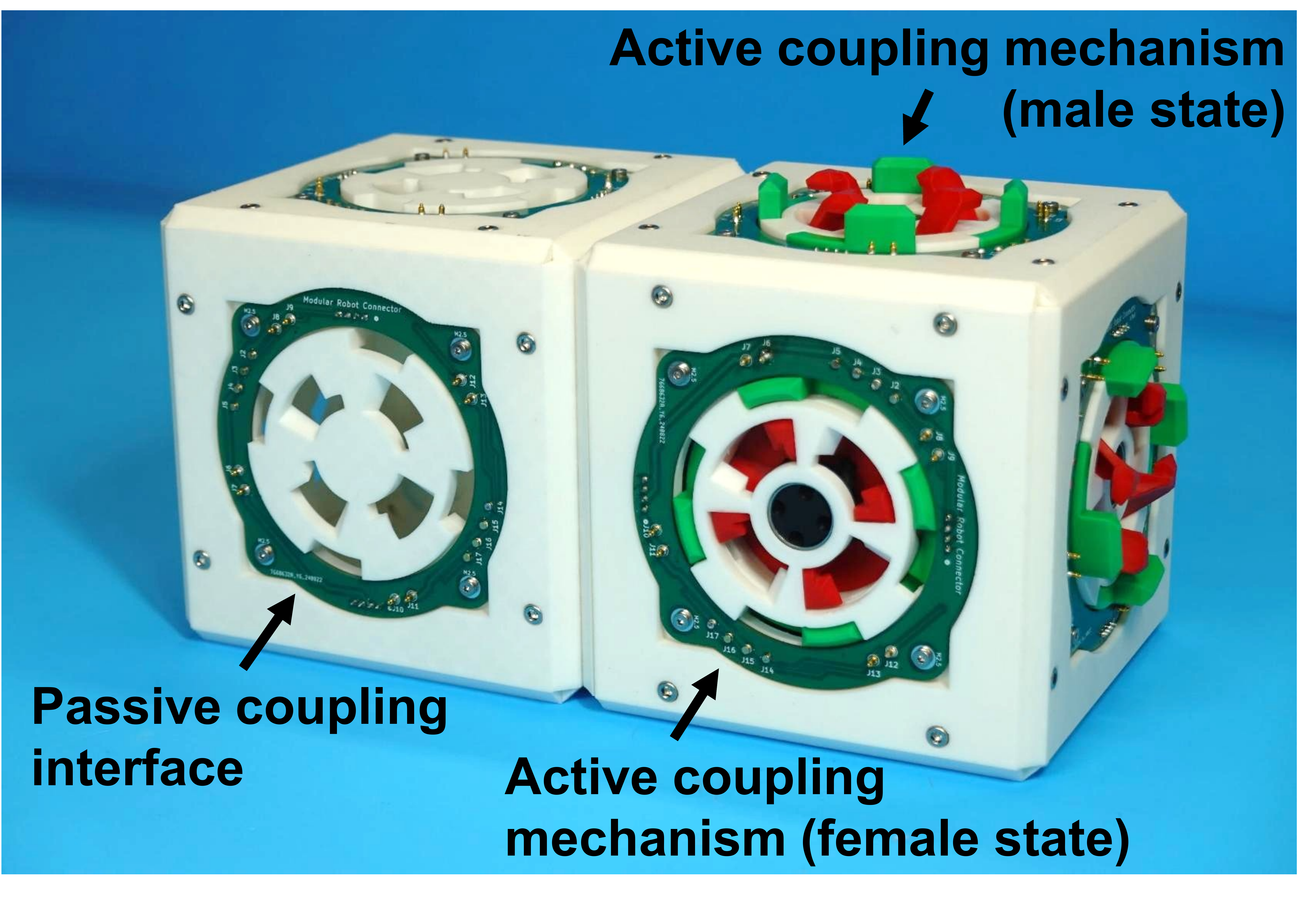}
    \caption{Heterogeneous structural modules with the proposed coupling mechanism.}
    \label{fig_connector_active}
\end{figure}

\section{Introduction} \label{intro}
Modular robots are robotic systems consisting of multiple independent modules coupled via coupling interfaces.
These systems can be reconfigured to create any assembly that meets the requirements of each environment or task\cite{mdl_review}, while also offering the flexibility to replace damaged modules.
Additionally, the modularity enables transportation in small units, making them particularly suitable for applications with strict size constraints, such as space missions.

To enable large-scale construction in extreme environments, including the lunar surface and in orbit, several researchers have proposed lattice-based modular robot systems that combine structural modules with assembler robots\cite{Terada2005-zv, MIT_asm_robot, ARMADAS}.
Modules equipped with active coupling mechanisms enable more flexible reconfiguration and can construct stronger structures compared to those using passive connection methods such as magnets.

A reliable coupling mechanism is a major challenge for modular robots, including lattice-based systems \cite{review_coupling}.
Coupling mechanisms must ensure secure connections, withstand external forces, and allow for precise alignment during assembly.
For lattice-based systems, additional requirements include maintaining flat surfaces when uncoupled, as modules connected on multiple faces must be able to move away horizontally without being blocked by protruding mechanisms on other faces.

\begin{table*}[t]
\begin{center}
\caption{Comparison of coupling mechanism}\label{tab:mechanisms_comparison}
\begin{tabular}{lcccccc}
\hline
\begin{tabular}[c]{@{}l@{}}Coupling mechanism /\\Robot platform\end{tabular} &
  Category &
  Gender &
  \begin{tabular}[c]{@{}c@{}}Single-sided\\ coupling\end{tabular} &
  \begin{tabular}[c]{@{}c@{}}Single-sided\\ decoupling\end{tabular} &
  \begin{tabular}[c]{@{}c@{}}Coupling to passive\\ coupling interface\\(in lattice module)\end{tabular} &
  \begin{tabular}[c]{@{}c@{}}Flat surface\\ (when not coupled)\end{tabular} \\ 
  \hline
  \hline
M-TRAN I\hspace{-1.2pt}I\hspace{-1.2pt}I\cite{M-TRAN} & Hooks       & Gendered   & No           & No           & \textbf{Yes} & \textbf{Yes} \\ 
\hline
Roombots\cite{Roombots} & Hooks       & Simultaneous Hermaphrodite & \textbf{Yes} & No           & \textbf{Yes} & \textbf{Yes} \\
HerCel  \cite{HerCel}   & Hooks       & Simultaneous Hermaphrodite & \textbf{Yes} & No           & \textbf{Yes} & \textbf{Yes} \\
CoBoLD  \cite{CoBoLD}   & Hooks       & Simultaneous Hermaphrodite & \textbf{Yes} & No           & \textbf{Yes} & \textbf{Yes} \\
  \hline
RoGenSiD\cite{RoGenSiD} & Shape Match & Genderless & No           & \textbf{Yes} & No & No           \\
HiGen   \cite{HiGen}    & Shape Match & Genderless & No           & \textbf{Yes} & No           & \textbf{Yes} \\
  \hline
SINGO   \cite{SINGO}    & Shape Match & Sequential Hermaphrodite & \textbf{Yes} & \textbf{Yes} & \textbf{Yes} & No           \\
GHEFT\cite{GHEFT_janl} & Shape Match & Sequential Hermaphrodite & \textbf{Yes} & \textbf{Yes} & \textbf{Yes} & No           \\
\textbf{Proposed mechanism} & Shape Match & Sequential Hermaphrodite & \textbf{Yes} & \textbf{Yes} & \textbf{Yes} & \textbf{Yes} \\ \hline
\end{tabular}
\end{center}
\end{table*}

Equipping all modules with active coupling mechanisms on all faces increases mechanical complexity and power consumption.
Castano and Will characterized the fundamental design tradeoff between homogeneous and heterogeneous systems: homogeneous systems increase hardware complexity by adding functionality to each module, while heterogeneous systems may reduce redundancy but allow for specialized modules\cite{hetero_first}.
Therefore, heterogeneous modular robotic systems are being explored, where some modules have active coupling mechanisms on specific faces, while the remaining faces are simplified to passive coupling interfaces without any actuators\cite{current_trends_with_hetero}.
This heterogeneous approach not only optimizes weight and power consumption but also reduces cost by installing expensive components such as sensors and actuators only where needed.
The approach requires active coupling mechanisms to be compatible with passive coupling interfaces in the same plane, which has received little attention in previous coupling mechanism designs.

We present a novel mechanical coupling mechanism designed for lattice-based modular robotic systems, particularly targeting space construction applications that combine structural modules with assembler robots (Fig.~\ref{fig_connector_active}).
The main contribution of this paper is the development of a coupling mechanism that enables flexible reconfiguration of heterogeneous lattice-based modular robots through its ability to actively transition between male and female states.

A key requirement for heterogeneous lattice-based systems is maintaining flat surfaces when uncoupled, and active coupling mechanisms must be capable of connecting with passive coupling interfaces on the same plane.
The proposed mechanism satisfies these requirements.

Another significant feature of the proposed mechanism is its ability to achieve single-sided coupling and decoupling.
This capability is critical for ensuring fault tolerance in modular systems.
Single-sided coupling and decoupling refer to the ability of one active coupling mechanism to perform the entire coupling or decoupling process, even if the opposing module is unpowered or malfunctioning. By enabling one-sided control, this mechanism supports flexible reconfiguration and enhances system reliability.

The mechanism significantly expands the design space for heterogeneous modular systems, enabling specialized modules to focus on specific functions while ensuring system-wide compatibility.
This approach opens new possibilities for space construction systems where resource efficiency and reliable operation are crucial.

\section{Related Work}\label{related_work}
In previous studies, coupling mechanisms for modular robots have often been classified based on their sexual characteristics.
This classification approach has been widely utilized but remains insufficient to meet the requirements of heterogeneous lattice-based modular robotic systems.

To address the limitations of existing classifications, we propose a biologically inspired reclassification.
In this framework, active coupling mechanisms with two distinct types or states are defined as "male" when they primarily feature convex elements and "female" when they primarily feature concave elements.

This reclassification introduces two new categories: \textit{Simultaneous Hermaphrodites} and \textit{Sequential Hermaphrodites}, inspired by biological nomenclature.
By adopting this terminology, we aim to provide a clearer and more intuitive framework for understanding the behavior and functionality of modular robot coupling mechanisms.

The following sections detail these categories, discussing their characteristics, strengths, and limitations in the context of heterogeneous lattice-based modular robots.
The features of the coupling mechanisms are summarized in Table \ref{tab:mechanisms_comparison}.

\subsection{Gendered}\label{gendered}
Gendered coupling mechanisms are characterized by their distinct male and female components, where male active mechanisms couple exclusively with female passive interfaces.
This inherent design restriction means that male-to-male or female-to-female connections are impossible.
This approach ensures compatibility between active mechanisms and passive interfaces but limits reconfiguration flexibility.
Specifically, while the male (active) side can initiate decoupling, the female (passive) side cannot, making single-sided decoupling impossible from the passive side.

M-TRAN I\hspace{-1.2pt}I\hspace{-1.2pt}I\cite{M-TRAN} is a typical this type mechanism, coupled by hooking the male hook of an active connector into the female hole of a passive interface.

\subsection{Simultaneous Hermaphrodite}
Simultaneous hermaphrodite coupling mechanisms, previously termed as bi-gendered or hermaphrodite mechanisms, feature both male and female elements that are permanently present and active simultaneously on each coupling interface.
During coupling, each mechanism's male elements engage with the other's female elements simultaneously, creating a symmetrical connection.

Roombots\cite{Roombots}, HerCel\cite{HerCel}, and CoBoLD\cite{CoBoLD} are these types.
These mechanisms typically activate four hooks as male elements and four corresponding holes as female elements on each interface.
This type has a significant limitation in terms of reconfiguration flexibility because single-sided decoupling is impossible.
Even when one mechanism's male elements are actuated, successful decoupling requires active participation from both coupling mechanisms.
This requirement makes the system less robust to failures, as a malfunctioning module cannot be removed by its neighboring modules.

\subsection{Genderless}
Genderless coupling mechanisms use identical coupling interfaces that connect without distinguishing between male and female roles.
Instead of using different types of elements, these mechanisms achieve coupling through synchronized actuation of identical components.

RoGenSiD\cite{RoGenSiD} and HiGen\cite{HiGen} are these types, where the rotating plates facing each other perform relative motion for coupling.
If one plate rotates in the opposite direction, single-sided decoupling is possible.
To be compatible with these mechanisms, passive coupling interfaces would need to maintain their coupling elements in the same state as active mechanisms after coupling actuation, making it impossible to achieve flat surfaces when uncoupled.
Consequently, these mechanisms are unsuitable for heterogeneous lattice-based modular robot systems.

\subsection{Sequential Hermaphrodites}
Sequential Hermaphrodite coupling mechanisms, although often classified as genderless in previous literature, represent an active system capable of switching between male and female states.
The two coupling mechanisms are mediated in advance, ensuring that one mechanism adopts the male state while the other assumes the female state, thereby facilitating successful coupling.

SINGO\cite{SINGO} and GHEFT\cite{GHEFT_janl} are these types.
The chuck-like arrangement of hooks can move to change into the male or female mode.
Due to the hooks' sliding movement, single-sided coupling and decoupling is possible from either the male or the female state side.
However, their hooks permanently protrude from the coupling surface.
This characteristic conflicts with a fundamental requirement of lattice-based systems where modules must maintain flat surfaces when uncoupled.

\begin{figure}[t]
    \centering
    \includegraphics[width=.85\linewidth]{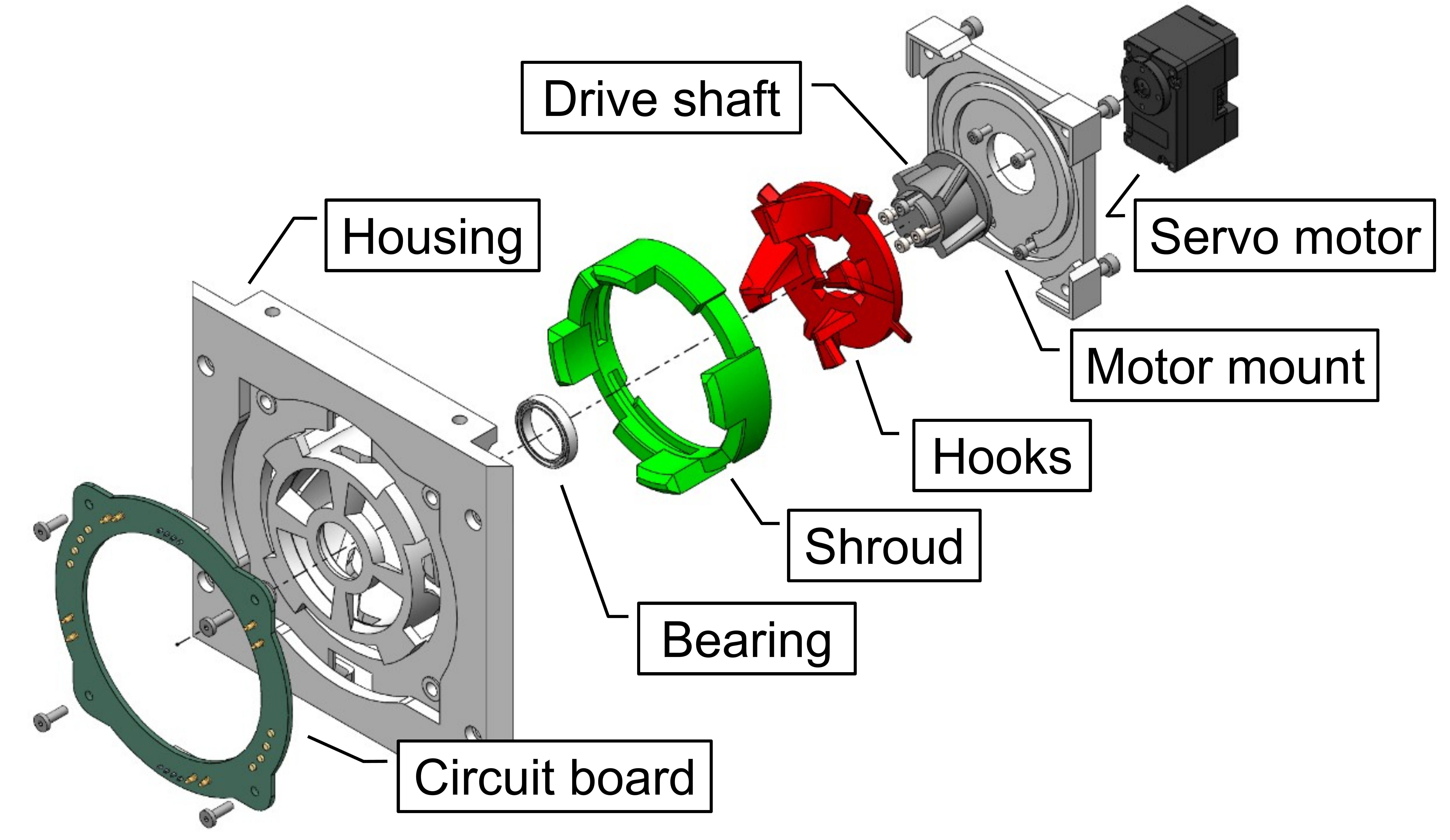}
    \caption{Exploded view of the proposed coupling mechanism.}
    \label{fig_Exploded View}
\end{figure}

\begin{figure}[t]
    \centering
    \includegraphics[scale=0.075]{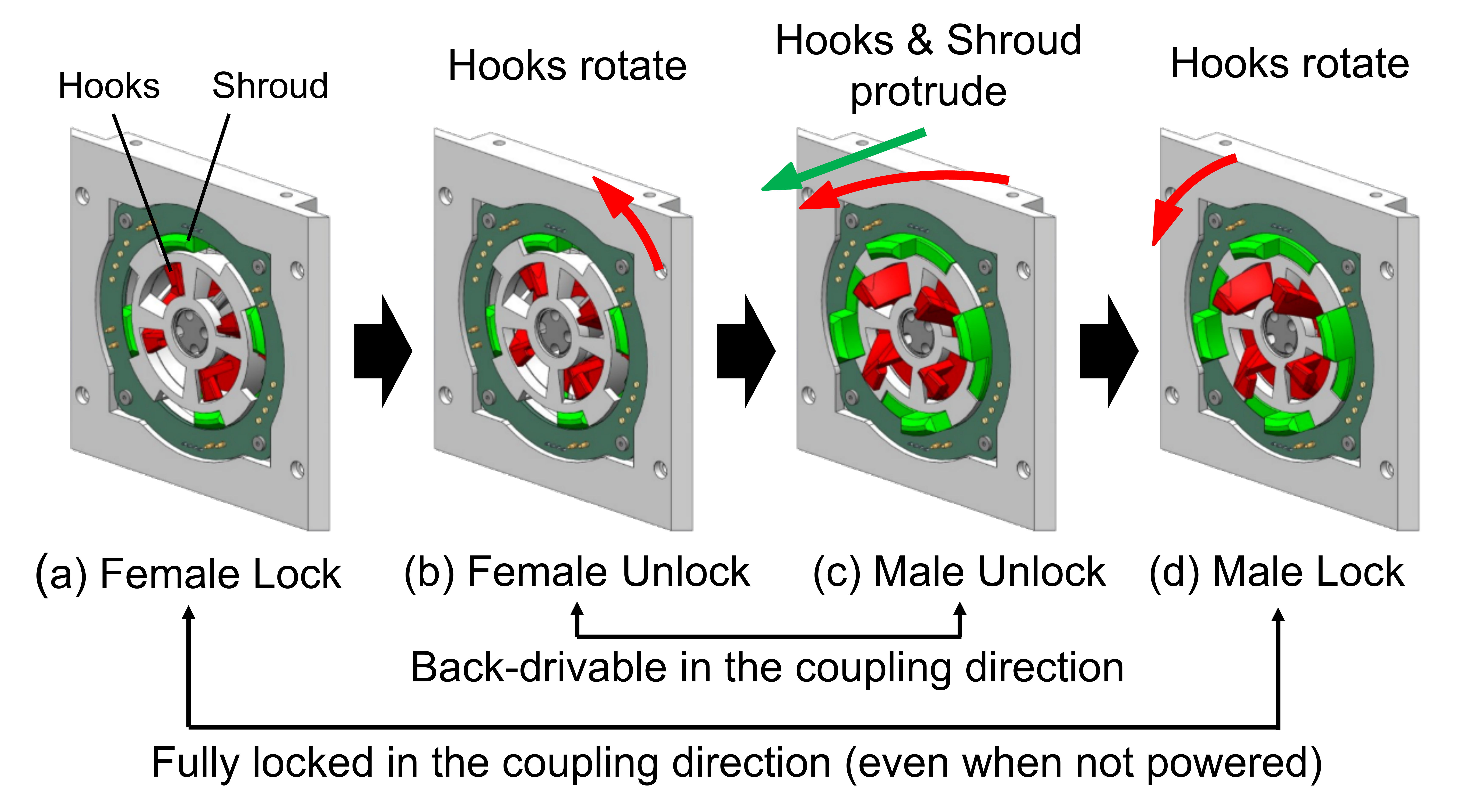}
    \caption{State transitions of the proposed coupling mechanism in its (a) Female Lock, (b) Female Unlock, (c) Male Unlock, and (d) Male Lock states.}
    \label{fig_all_gender}
\end{figure}

\begin{figure*}[thb]
    \centering
    \includegraphics[scale=0.076]{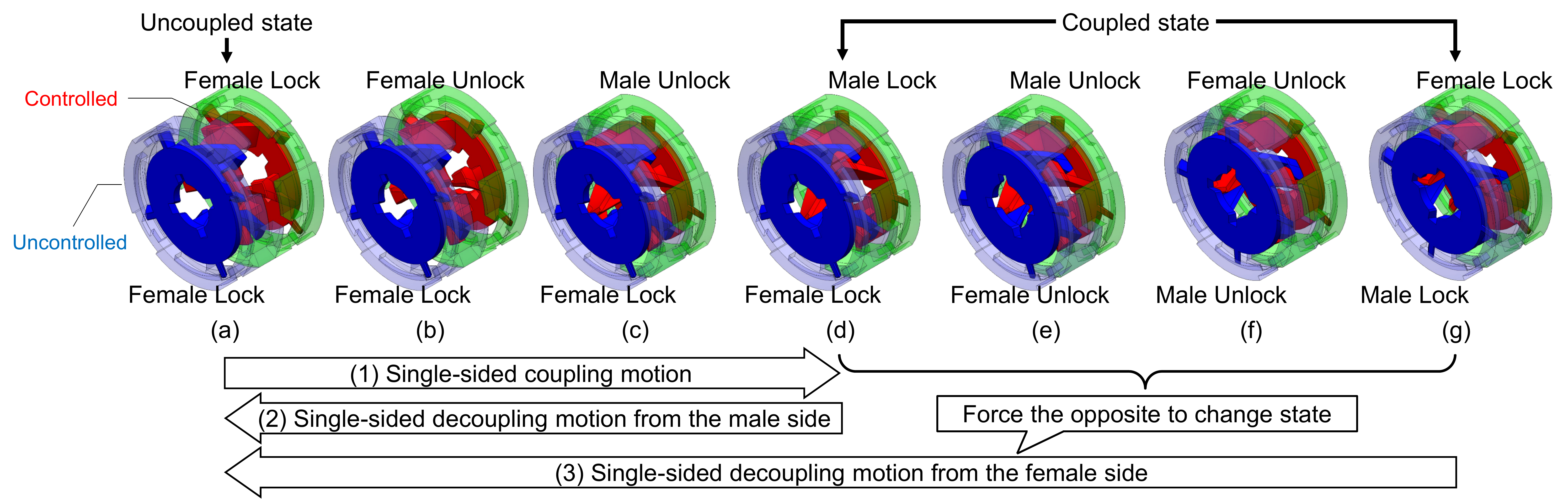}
    \caption{State transitions of the proposed coupling mechanism during (1) Single-sided coupling motion, (2) Single-sided decoupling motion from the male side, and (3) Single-sided decoupling motion from the female side.
    }
    \label{fig_coupling_decoupling}
\end{figure*}

\section{Methodology}\label{methodology}
We propose a novel mechanical coupling mechanism for a heterogeneous lattice-based modular robot system.
We selected the Sequential Hermaphrodite type as the characteristic of the gender.
This is because the type is the only gender that can satisfy the requirements that the active coupling mechanism can couple with the passive coupling interface, and that single-sided coupling and decoupling are possible.
In addition, unlike existing Sequential Hermaphrodite type coupling mechanisms, the coupling mechanism can be flat when not coupled, so it can be used in lattice-based modular robots.

This mechanism is designed as one side of a cubic lattice module and 90-degree symmetry.
As shown in Fig. \ref{fig_Exploded View}, it consists of 3D printed PLA plastic components, a bearing, a contact circuit board, and a servo motor.
The servo motor is the Dynamixel XC330-T288-T provided by ROBOTIS.
The lattice module, which is equipped with the proposed coupling mechanism on all sides, is \SI{120}{mm} x \SI{120}{mm} x \SI{120}{mm} in size and weighs \SI{840}{g}.
The plastic components consist of a housing, hooks, a shroud, a drive shaft, and a motor mount.
The protrusions on the drive shaft engage the inner groove of the hooks, and the hooks move with the drive shaft.
The projection of the drive shaft extends from the side of the servo motor in a clockwise spiral.
When the drive shaft rotates counterclockwise, the force vector applied to the hooks has both counterclockwise and axial forward components.
The housing has a through groove that engages the outer projection of the hook, and the hooks moves by following the groove.
From these elements, the simple rotational motion of the servo motor is converted into a complex motion that combines the rotational and linear motion of the hooks.
The shroud has grooves that engage with the outer grooves of the hooks only in the circumferential direction, and the shroud follows only the linear motion of the hook.

As shown in Fig.~\ref{fig_all_gender}, the grooves in the housing enable the hook to transition to four states.
When uncoupled, all modules are in state (a) Female Lock.
From (a) to (b) Female Unlock, hooks move only in the circumferential direction and not in the axial direction and the shroud does not move.
From (a) to (b), the hooks and shroud are housed and a flat plane is realized.
From (b) to (c) Male Unlock, the hooks move not only in the circumferential direction but also in the axial direction and protrude outward from the plane of the module.
From (c) to (d) Male Lock, the hooks rotate only circumferentially again and do not move axially and the shroud does not move.

The behavior of the proposed coupling mechanism is inspired by and has characteristics similar to those of Genderless coupling mechanisms such as RoGenSiD\cite{RoGenSiD} and HiGen\cite{HiGen}.
In the Genderless coupling mechanisms, the coupling can be achieved only when both coupling mechanisms facing each other are controlled and hooks are extended.
The coupling plane is the middle plane of the two modules.

On the other hand, the proposed mechanism is of the Sequential Hermaphrodites type, in which a module in the Male Lock state and a module in the Female Lock state are coupled.
As a result, single-sided coupling is possible and has the advantage of connecting to a module that is not supplied with power or to a module that is broken.
In addition, the coupling plane exists within the module that is in the Female Lock state.
This allows coupling to passive interfaces that do not have actuators.

\begin{figure}[h]
  \begin{minipage}{1.0\hsize}
    \centering
    \includegraphics[keepaspectratio, scale=0.09]{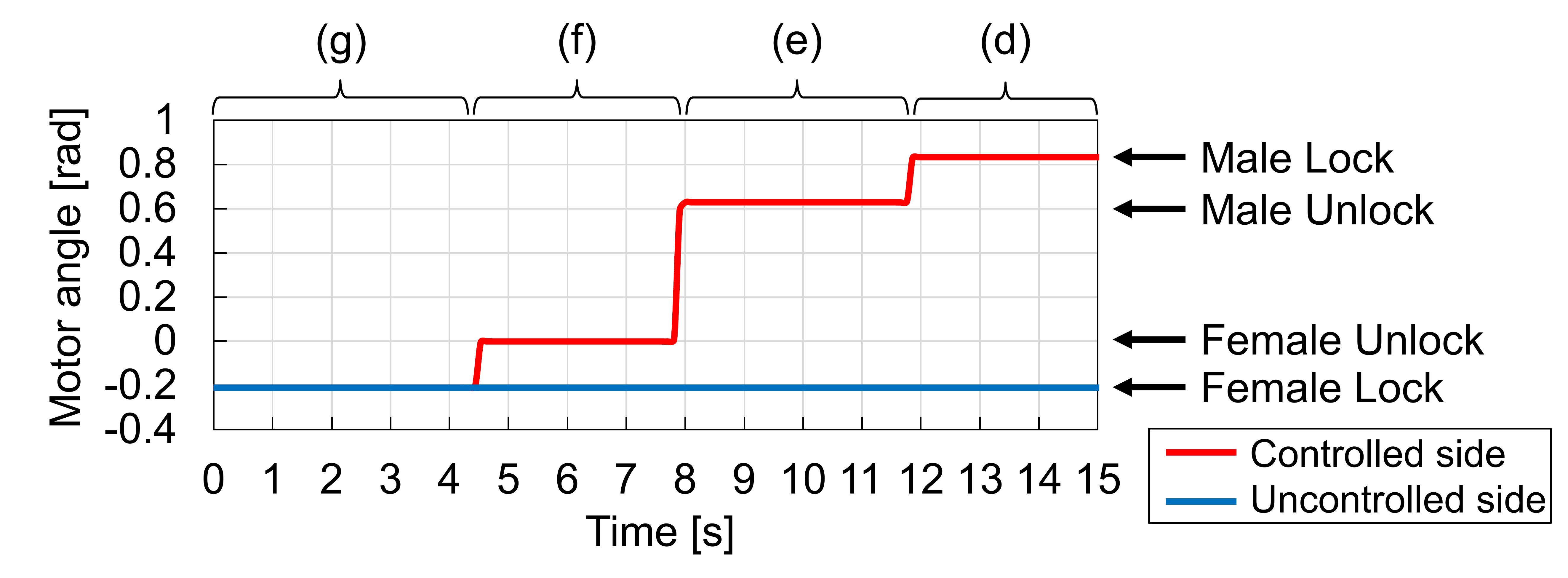}
    \subcaption{Single-sided coupling motion.}
    \label{fig_motion_coupling}
  \end{minipage}\\
  \begin{minipage}{1.0\hsize}
    \centering
    \includegraphics[keepaspectratio, scale=0.09]{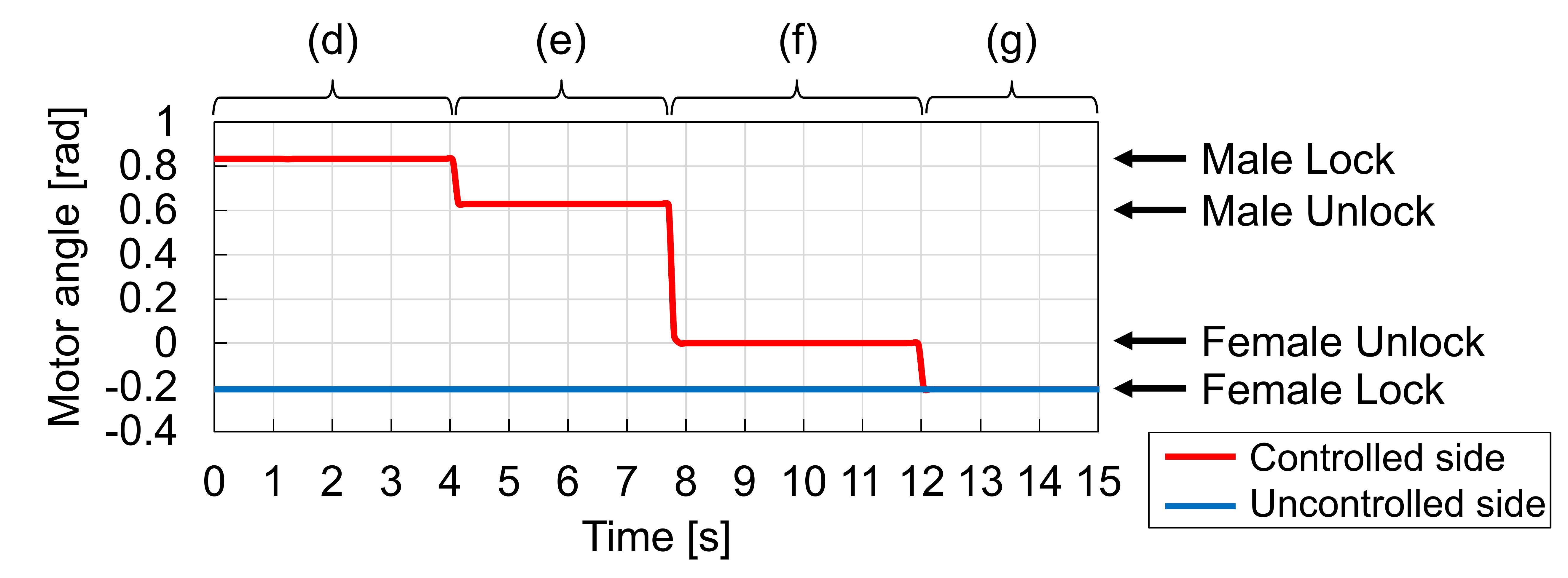}
    \subcaption{Single-sided decoupling motion from the male side.}
    \label{fig_motion_decoupling_male}
  \end{minipage}\\
  \begin{minipage}{1.0\hsize}
    \centering
    \includegraphics[keepaspectratio, scale=0.09]{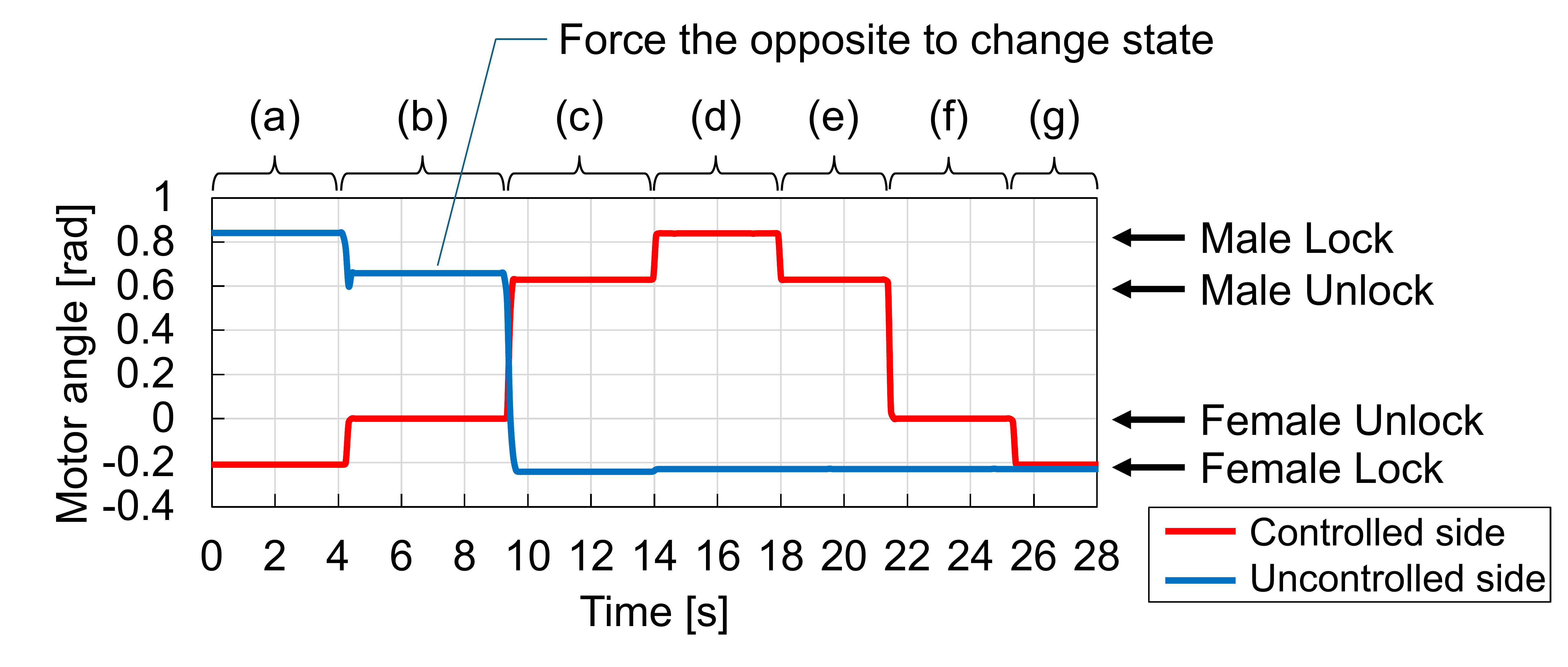}
    \subcaption{Single-sided decoupling motion from the female side.}
    \label{fig_motion_coupling_female}
  \end{minipage}
\caption{
Results of single-sided coupling/decoupling tests.
The graphs show the servo motor angle changes during the motion.
The ranges labeled as (a)–(g) correspond to the respective states shown in Fig.~\ref{fig_coupling_decoupling}.
}\label{3fig}
\end{figure}

Fig. \ref{fig_coupling_decoupling} shows the state transitions of coupling/decoupling.
In all states, red modules are controlled, and blue modules are not.
First, during (1) single-sided coupling, the transitions are (a), (b), (c), and (d), in that order.
In state (d), the hooks engage each other to constrain axial movement.
The shroud also engages to constrain movement in the direction parallel to the coupling plane.
Therefore, even if an external force is applied, the force to rotate the hooks does not work, and the coupling can be maintained even when the power supply to the servo motors of both parts is stopped.
The hooks and shrouds are made of elements spaced 90 degrees apart, which makes the coupling 90 degrees rotationally symmetrical.
In addition, they are filleted to allow for misalignment in position and angle.

When coupled, the modules are electrically connected to each other by spring-loaded pins mounted on the circuit board.
Considering the 90-degree rotational symmetry of the connectors, a staggered placement similar to that of HiGen\cite{HiGen} was adopted.

Second, at the time of (2) single-sided decoupling from the male side, the state transition should be reversed from the case of coupling, in the order of (d), (c), (b), and (a).

Third, at the time of (3) single-sided decoupling from the female side, transition the states in the order of (g), (f), (e), (d), (c), (b), and (a).
In (g), no power is applied to the actuator of the module on the Male Lock side, and it can be back-driven by the module on the Female Lock side while engaged.
Using this, the state of the other side is forced to change from Male Lock to Female Lock.
After that, since the states of Male and Female are reversed, they can be decoupled by the same action as when single-sided decoupling is performed from the Male side.

\section{Experiments and Discussions}\label{results_discussions}


\subsection{Single-Sided Coupling/Decoupling Tests}\label{exp_coupling_decoupling_motion}
The single-sided coupling/decoupling motions of the proposed mechanism were conducted and the encoder values of the servo motors were plotted during the motions.
In order to make it easier to show the coupling, a delay is included between each state.
Fig.~\ref{fig_motion_coupling} shows the single-sided coupling, Fig. \ref{fig_motion_decoupling_male} shows the single-sided decoupling from the male side, and Fig. \ref{fig_motion_coupling_female} shows the single-sided decoupling from the female side.
Each of these shows the same behavior as described in Sect.~\ref{methodology}.
In particular, Fig.~\ref{fig_motion_coupling_female} shows that the controlled coupling mechanism side forces the uncontrolled coupling mechanism to change its state.
They also show that the motion time is fast.

\begin{figure}[t]
    \centering
    \includegraphics[scale=0.065]{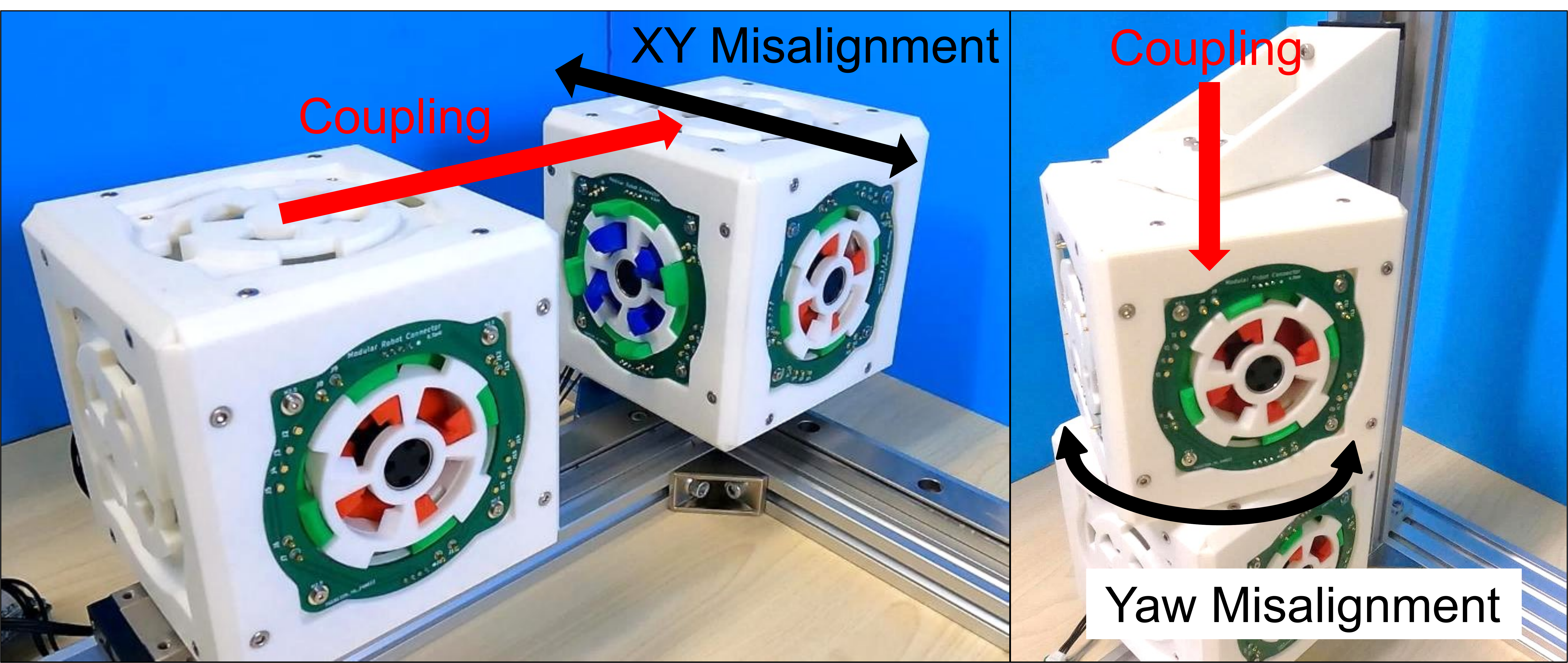}
    \caption{Experimental setup for Misalignment Tolerance Tests. Displacement misalignment in the X-Y direction and angular misalignment in yaw rotation were evaluated.}
    \label{fig_env_experiment}
\end{figure}

\begin{figure}[h]
  \begin{minipage}{1.0\hsize}
    \centering
    \includegraphics[keepaspectratio, height=.7\linewidth]{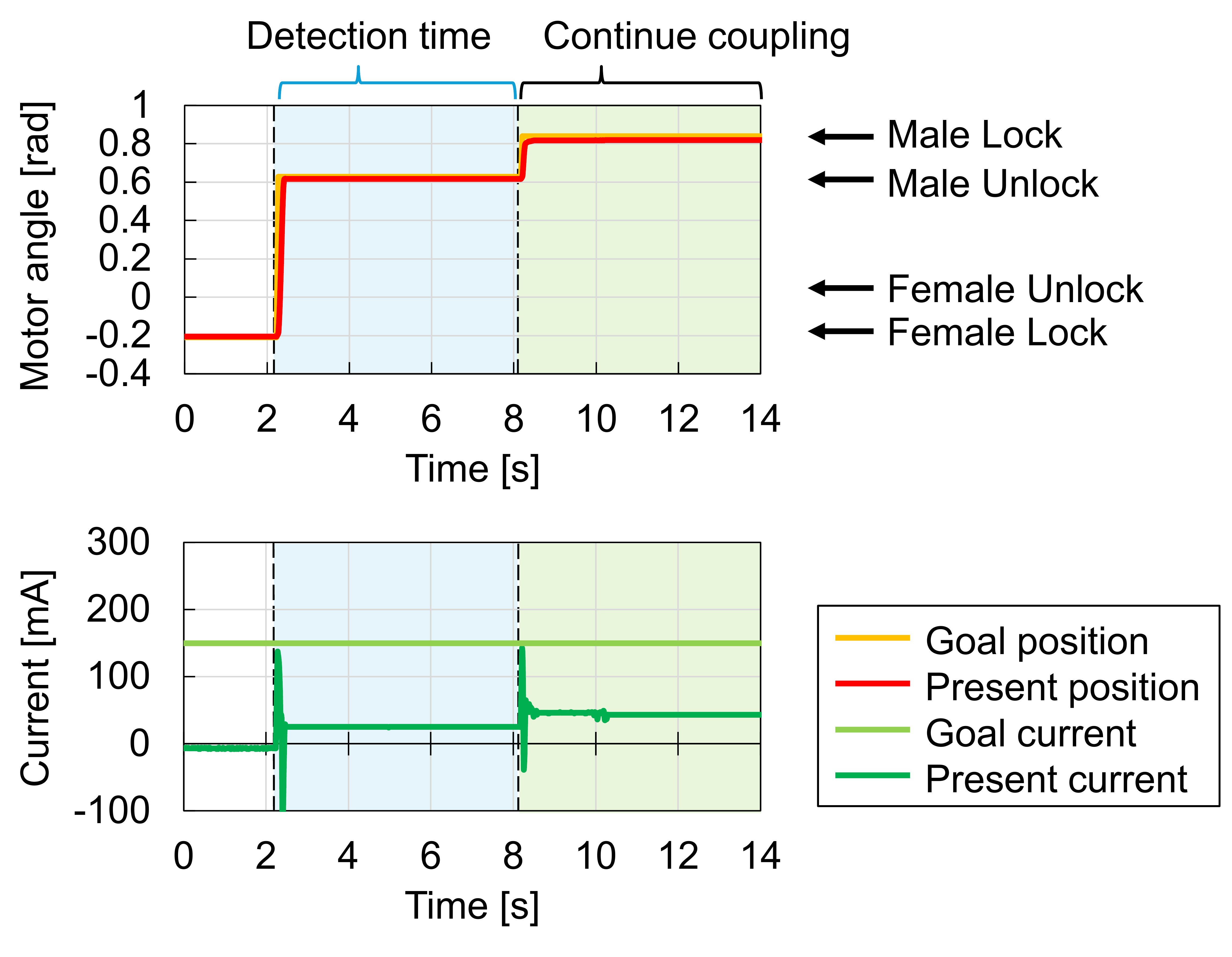}
    \subcaption{Transition of the servo motor angle and current during a successful coupling.}
    \label{fig:pos_current_success}
  \end{minipage}\\
  \begin{minipage}{1.0\hsize}
    \centering
    \includegraphics[keepaspectratio, height=.7\linewidth]{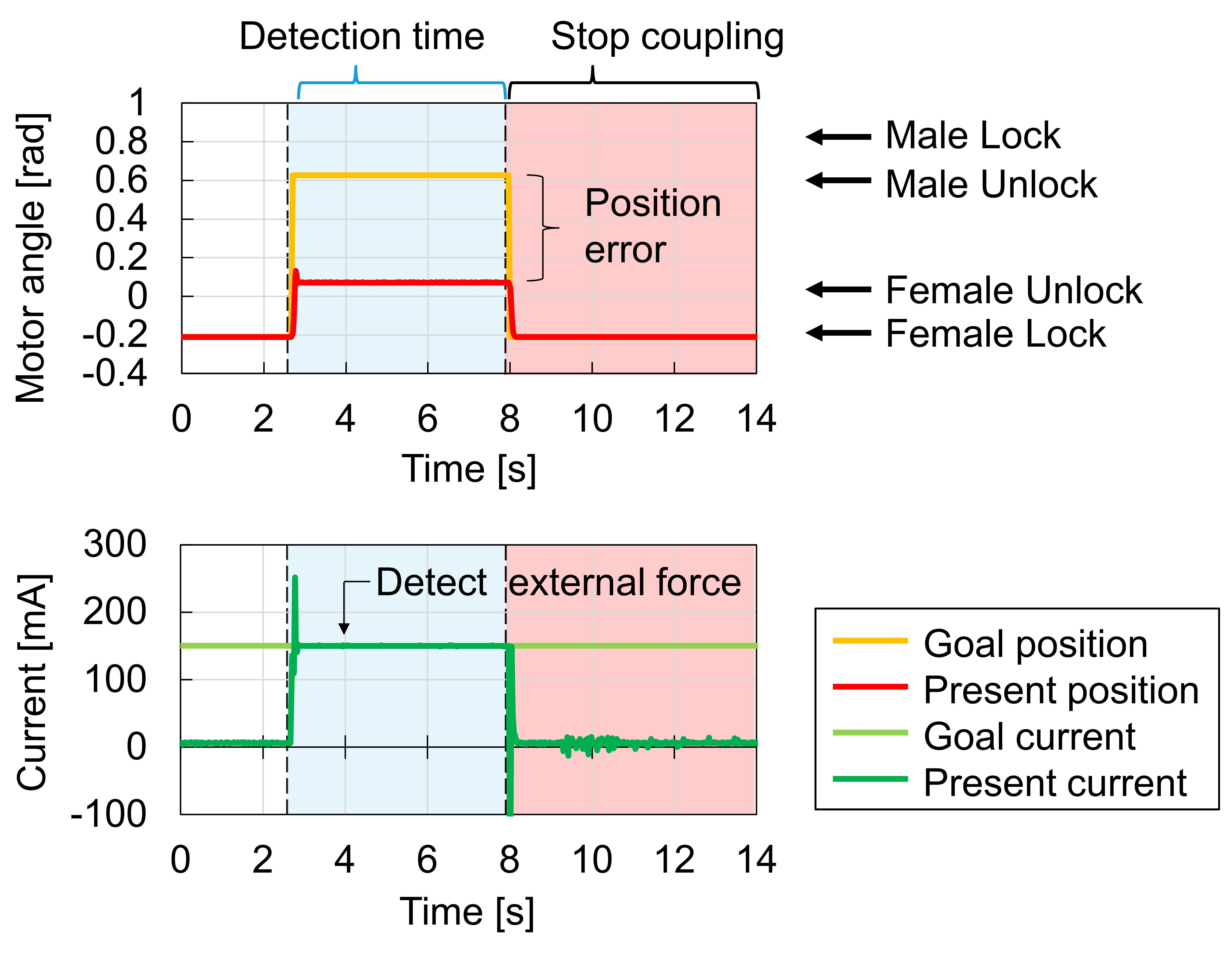}
    \subcaption{Transition of the servo motor angle and current during a failed coupling.}
    \label{fig:pos_current_fail}
  \end{minipage}\\
\caption{
Servo motor angle and current profiles during Fail-Safe System Tests. Fig.~\ref{fig:pos_current_fail} shows the case where the coupling target is not in the correct position, and the mechanism stopped to couple. Fig.~\ref{fig:pos_current_success} shows the case where the coupling target is correctly positioned, resulting in successful coupling.}
\label{fig:pos_current_demo}
\end{figure}

\subsection{Misalignment Tolerance Tests}\label{exp_mesalignment}
Experiments were conducted to test the tolerance of the proposed coupling mechanism for position and angle misalignment of the coupling.
The surface of the housing of the mechanism is the X-Y plane and the coupling direction is the Z axis.
The mechanism assumes that modules are coupled with each other by an assembly robot arm.
When coupling two modules, the robot arm is controlled to push the surfaces of the two modules against each other by admittance control or force control using a force sensor mounted on the arm\cite{Luca}.
This can almost minimize the position error in the Z axis direction and the angular error in the X axis rotation (roll) and Y axis rotation (pitch) directions.
On the other hand, the position error in the X or Y axis direction and the angular error in the Z-axis rotation (yaw) direction must be compensated for by the coupling mechanism.
Therefore, the tolerances of the position error in the X and Y axes direction and the angular error in the yaw rotation direction were tested.
Using a linear motion guide, movement in directions except the direction to be tested was constrained.
The experiment environment is shown in Fig.~\ref{fig_env_experiment} and the result is shown in Table~\ref{tab:misapignment_load}.
Unacceptable misalignment can be tolerated by the fail-safe system shown in Sect.~\ref{exp_fail_system}.

\begin{table}[]
\centering
\caption{Misalignment tolerance and load capability}
\label{tab:misapignment_load}
\begin{tabular}{ll}
    \hline
    \begin{tabular}[c]{@{}l@{}}Displacement misalignment of\\ X and Y axes\end{tabular}
    & \begin{tabular}[c]{@{}l@{}} \SI{\pm 2.5}{mm}\\(2.9\% of the circuit size)\end{tabular}\\
    \hline
    \begin{tabular}[c]{@{}l@{}}Angle misalignment of Yaw axis\end{tabular}
    & $\pm$ $\ang{2.5}$ \\
    \hline
    Load capability of X and Y axes
    & $\ge$ \SI{300}{N}\\
    \hline
    Load capability of Z axis & \SI{129}{N}\\
    \hline
\end{tabular}
\end{table}

\subsection{Load Capability Tests}\label{exp_load}
To indicate that the coupling mechanism has sufficient coupling strength even in the non-powered state, load capacity tests were conducted.
After coupling the two coupling mechanisms, the power supply was shut off.
Following that, measurements were conducted by pulling a wire from the coupling direction (Z axis direction) and from a direction parallel to the coupling plane (X and Y axes direction) with a force gauge.
The result is shown in Table~\ref{tab:misapignment_load}.
Although the load capacity is already sufficient, it can be increased by changing the material if more load capacity is required.

\subsection{
Fail-Safe System Tests
}\label{exp_fail_system}
The servo motor in the proposed coupling mechanism can measure the current value, from which the approximate torque can be estimated.
In the process from Female Unlock to Male Unlock, the goal current is set low (\SI{150}{mA}) for position control.
This control enables the hooks and shroud to softly protrude in the coupling direction and detect external force during the process.
When the coupling is successful, as shown in Fig.~\ref{fig:pos_current_success}, the servo motor checks to the Male Unlock position, then rotates to Male Lock and couples.
Therefore, when the coupling fails as shown in Fig.~\ref{fig:pos_current_fail}, the failure can be detected by monitoring the current and preventing the coupling mechanism from being broken by forcibly exerting torque on the servo motor.

\section{Conclusions}\label{conclusions}
In this paper, we proposed a novel mechanical coupling mechanism for lattice-based modular robotic systems.
The proposed coupling mechanism has 90-degree symmetry and can realize electrical connections for communication and power supply, and the mechanical hooks can keep the coupling even when the power is not supplied.
It is capable of single-sided coupling and decoupling.
It can couple to the passive interface in the same plane as when two mechanisms are coupled to each other.
The surface of the mechanism can be flat when uncoupled, which allows for lattice-based structures.
The proposed coupling mechanism is the first coupling mechanism with these features.
We built the coupling mechanism and tested its single-sided coupling and decoupling motion, tolerance for misalignment, and loading capacity.
We also showed that the coupling mechanism can be used as fail-safe system by monitoring the current.
Future work will include the completion of a system in which heterogeneous structural modules using this coupling mechanism are assembled by an assembler robot.

\section*{Acknowledgment}

This work was supported by JST Moonshot R\&D Program, Grant Number
JPMJMS223B.
The authors also thank Prof. Fumitoshi Matsuno's research group for the invaluable discussions with them in Moonshot R\&D Program.

\bibliographystyle{IEEEtran}
\bibliography{reference}

\end{document}